\pdfoutput=1

\documentclass[11pt]{article}

\usepackage{acl}

\usepackage{times}
\usepackage{latexsym}

\usepackage[T1]{fontenc}

\usepackage[utf8]{inputenc}
\usepackage{CJKutf8}

\usepackage{microtype}

\usepackage{booktabs, multirow} 
\usepackage{soul}
\usepackage{changepage,threeparttable} 

\usepackage{amsfonts, amsmath, amssymb}
\usepackage{xcolor}
\usepackage{calc}
\usepackage{ifthen}
\usepackage{tikz}
\usepackage{url} 
\usepackage{tabularx}
\usepackage{float}
\usepackage{rotating,booktabs,multirow}
\usepackage{adjustbox}
\usepackage{xcolor}

\usepackage{makecell}

\usepackage[caption=false]{subfig}

\title{Exploring and Adapting Chinese GPT to  Pinyin Input Method}

\author{
Minghuan Tan$^1$\thanks{~~Work done during internship at Tencent AI Lab. ~~~~* indicates equal contribution.} \and
Yong Dai$^{2*}$\and
Duyu Tang$^2$\thanks{~~Corresponding author.}\and
Zhangyin Feng$^2$ \\
\bf{Guoping Huang}$^2$ \and 
\bf{Jing Jiang}$^1$ \and
Jiwei Li$^3$  \and
Shuming Shi$^2$
\\ \\
$^1$ Singapore Management University.
$^2$ Tencent AI Lab.
$^3$  Zhejiang University.\\
mhtan.2017@phdcs.smu.edu.sg, jingjiang@smu.edu.sg \\
\{yongdai,duyutang,aifeng,donkeyhuang,shumingshi\}@tencent.com,  
jiwei\_li@zju.edu.cn
}

\begin{document}

\maketitle
\begin{CJK*}{UTF8}{gbsn}

\begin{abstract}
While GPT has become the de-facto method for text generation tasks, its application to pinyin input method remains unexplored.
In this work, we make the first exploration to leverage Chinese GPT for pinyin input method.
We find that a frozen GPT achieves state-of-the-art performance on perfect pinyin.
However, the performance drops dramatically when the input includes abbreviated pinyin.
A reason is that an abbreviated pinyin can be mapped to many perfect pinyin, which links to even larger amount of Chinese characters.
We mitigate this issue with two strategies,
including enriching the context with pinyin and optimizing the training process to help distinguish homophones. 
To further facilitate the evaluation of pinyin input method, we create a dataset consisting of 270K instances from 15 domains.
Results show that our approach improves the performance on abbreviated pinyin across all domains.
Model analysis demonstrates that both strategies
contribute to the performance boost.
\end{abstract}

\vspace{0.01cm}

\vspace{0.01cm}
\section{Introduction}
\label{sec:intro}
GPT~\cite{radford2018improving,radford2019language} is a Transformer-based \cite{vaswani2017attention} language model that predicts tokens in an autoregressive manner.
With a generic model architecture and the availability of vast web text data, GPT has been successfully developed for English, Chinese~\cite{GPT2-Chinese,cpm-v1}, and many other languages.
It shows extraordinary ability to generate fluent sentences and has been successfully applied to a wide range of natural language generation tasks.
However, it remains unexplored to what extent GPT handles Chinese pinyin input method\footnote{\url{https://en.wikipedia.org/wiki/Pinyin_input_method}}, 
which is used by hundreds of millions people when they enter Chinese characters on computers and cellphones.

Pinyin input method allows users to enter Chinese characters based on their pronunciations. 
Given a pinyin\footnote{\url{https://en.wikipedia.org/wiki/Pinyin}} as the input, pinyin input method returns a list of Chinese characters pronounced with that pinyin.
Fundamental elements of pinyin include initials (声母) and finals (韵母). In most cases, a Chinese character is spelled with one initial followed by one final.
For example, as shown in Table~\ref{tab:intro-example}, the initial and final for the Chinese character ``我 \ (me)''  are \texttt{w} and \texttt{o}, respectively.  
\begin{table}[t]
\centering
\begin{tabular}{c|c|c|c}
\toprule
\textbf{ Character} & \textbf{Perfect Pinyin} &\textbf{Initial} & \textbf{Final}\\
\midrule
我& \texttt{wo} & \texttt{w} & \texttt{o} \\
们& \texttt{men} & \texttt{m} & \texttt{en} \\
\bottomrule
\end{tabular}
\caption{Examples of initials and finals for Chinese characters ``我们 \ (we)''.}
\label{tab:intro-example}
\end{table}
People may enter \textbf{perfect pinyin} (e.g., ``\texttt{wo men}'' for ``我们''), where initials and finals of all Chinese characters are entered. 
There are about 420 perfect pinyin in common use.
Sometimes, especially when multiple Chinese characters are entered at once, people may use \textbf{abbreviated pinyin} by only entering the initials of characters (e.g.,  ``\texttt{w m}'' for ``我们'').


\begin{table*}[ht]
\centering
\begin{tabular}{c|l|c|c|c}
\toprule
Id&\textbf{Context of Characters} & \textbf{Input Pinyin} & \textbf{Target} & \textbf{Pinyin Type}\\
\midrule
s1&我下周有时间，除了& \texttt{li bai yi you dian shi} & {礼拜一有点事} & Perfect \\
s2&我下周有时间，除了& \texttt{l b y y d s} & {礼拜一有点事} & Abbreviated\\
s3&老板帮我解决了难题，& \texttt{l b y y d s} & {老板永远滴神} & Abbreviated\\
\bottomrule
\end{tabular}
\caption{Illustrative examples of the task of pinyin input method with perfect pinyin and abbreviated pinyin. In s3, the input pinyin ``\texttt{l b y y d s}'' is the abbreviation of ``\texttt{lao ban yong yuan di shen}''. The translations of s1 and s3 are ``I am free next week except for the next Monday.'' and ``Boss helps me overcome the obstacle. You are the greatest of all time.'', respectively. }
\label{tab:task-definition-example}
\end{table*}
This work, to the best of our knowledge, is the first one to explore the use of Chinese GPT for pinyin input method. 
We start by testing the performance of a frozen GPT.
In this setting, we fix the parameters of GPT and predict Chinese characters from left to right in an autoregressive manner. 
At each time step, only characters pronounced with the same pinyin are legitimate candidates to be predicted.
We find that, when the input is perfect pinyin, a frozen GPT performs comparably to state-of-the-art systems on the benchmark dataset \cite{yang-etal-2012-towards}.
However, when the input is abbreviated pinyin with only initials of characters, the performance of GPT has a drastic drop.
A major reason is that an abbreviated pinyin maps to many perfect pinyin. 
For example, the initial ``\texttt{w}'' can be the abbreviation for ``\texttt{wo}'', ``\texttt{wei}'', ``\texttt{wang}'', ``\texttt{wai}'', ``\texttt{wu}'',  etc. 
This would lead to exponentially larger number of legitimate candidates of Chinese characters.
We mitigate this problem by incorporating pinyin information from two directions.
One is to enrich the input by adding pinyin as additional context.
The other is learning over pinyin-constrained vocabulary, which enhances the model's ability to distinguish between Chinese characters pronounced with the same pinyin. 

To further facilitate the research on pinyin input method, we construct a new dataset based on the WuDaoCorpora~\cite{YUAN202165}.
Our dataset includes 270K instances from 15 commonly used news domains.
To evaluate towards multiple facets, the dataset covers instances with different numbers of context characters and pinyin.
From our experiment results, we have these key findings:
\begin{enumerate}
    \item On {perfect pinyin}, frozen  GPT achieves state-of-the-art results. 
    \item On {abbreviated pinyin}, the performance of frozen GPT drops drastically. 
    Context enrichment with pinyin and  pinyin-constrained training both improve the performance.
    \item The performance of GPT-based models increases as the context of Chinese characters becomes longer.
\end{enumerate}


\section{Task}


The input of pinyin input method includes a sequence of Chinese characters $C=\{w_1,\dots,w_n\}$ as the context and a sequence of pinyin $P=\{p_{n+1},\dots,p_{n+k}\}$, where $w_i\in\mathcal{V}_w$, $p_{n+j}\in\mathcal{V}_p$, and $\mathcal{V}_w$ and $\mathcal{V}_p$ are the vocabularies of words and pinyin, respectively.
The output is a sequence of Chinese characters $O=\{w_{n+1},\dots,w_{n+k}\}$, where  $w_{n+i}\in\mathcal{V}_w$.
The number of output characters is the same as the number of pinyin (i.e., $k$) and each character should be pronounced with the corresponding pinyin. The output sequence is desired to follow the context of Chinese characters to form a coherent sentence. 
As mentioned earlier in the introduction section, the input pinyin might be perfect (e.g., ``\texttt{wo men}'') or abbreviated (e.g., ``\texttt{w m}''). Examples of the task are given in Table~\ref{tab:task-definition-example}.\footnote{People may also input pinyin like ``\texttt{l b y you dian shi}'', we leave this as a future work.}
In our definition, one situation is that the context of characters is empty, which corresponds to the scenario that people are entering pinyin at the beginning of a sentence.
The other situation is that the context includes real words, which stands for the scenario that people are entering pinyin in the middle of a written sentence. 

In this paper, we assume that the oracle pinyin segmentation results are provided.
Sometimes, a raw pinyin sequence can be mapped to different segmentation results.
For example, the raw pinyin input ``\texttt{jianshi}'' can be segmented as ``\texttt{ji an shi}'' (``集安市'', a city in the southwestern part of Jilin province, China) or ``\texttt{jian shi}'' (``见识'', which is translated as ``experience'' in English). 
Pinyin segmentation is a subtask \cite{zhao-etal-2006-improved,p2c-seg} of pinyin input method, which is well solved with the accuracy of 98\% \cite{zhang-etal-arxiv-tracing}.
We leave the integration of pinyin segmentation as future work.





\section{Models}
\label{sec:models}
In this section, we first introduce standard text-based GPT models adopted in this work (section \ref{sec:gpt}).
Afterwards, we introduce how to extend GPT models for pinyin input method with enriched pinyin context (section \ref{sec:pinyin-context}) and pinyin-constrained training (section \ref{sec:training}), respectively.


\subsection{GPT Baselines}
\label{sec:gpt}
In this work, we use character-level Chinese GPT as the backbone.
We describe character-level GPT models in this subsection.

We start with a publicly available character-level GPT  \cite{GPT2-Chinese}\footnote{\url{https://github.com/Morizeyao/GPT2-Chinese}}, which we call \textbf{GPT (public)}.
The model has the same configuration as the standard 12-layer GPT\footnote{\url{https://huggingface.co/gpt2}}. 
It is trained on the CLUECorpusSmall dataset of 14GB \cite{CLUECorpus2020}, which consists of Chinese news, Wikipedia, online forum message, and consumer comments.
We have tried another well known Chinese pretrained language model called CPM \citep{cpm-v1}, which is
trained on 100GB  data. The vocabulary of CPM contains both Chinese characters and words.\footnote{A Chinese word may consist of multiple Chinese characters. For example, the word ``我们'' (we) includes two characters ``我'' and ``们''. } 
We built a baseline with the CPM model of 12 layers\footnote{\url{https://github.com/TsinghuaAI/CPM-1-Distill}} and forced the generated token to be a Chinese character. 
However, this baseline does not work well on pinyin input method, partly because our character-level decoding is inconsistent with the way how CPM is trained.
It is promising to leverage the advantage of CPM on word-level decoding, and we leave this as a future work. 

To build a stronger Chinese GPT baseline, we use GPT (public) as the starting point and further pretrain on a 800GB data crawled by us  that is composed of news, Wikipedia, and novel texts. 
The model is trained with a batch size of 2,560 on 32x Tesla V100 GPUs.
We adopt the Adam optimizer~\cite{kingma2014adam} and set the learning rate to 1e-5 with a linear warmup scheduler. We run the warmup process for 10k steps and train 100k steps in total. 
We call this 12-layer GPT model as \textbf{GPT (ours)}.

To apply GPT (public) and GPT (ours) to pinyin input method, we use the traditional decoding pipeline of GPT to generate the sequence of Chinese characters in an autoregressive way. 
After encoding all the context of characters, the model predicts a Chinese character at each time step conditioned on the pinyin.
Only Chinese characters pronounced with the same pinyin are legitimate candidates to be predicted.
Without further clarification, this strategy is used in all the experiments.



\begin{figure*}[t]
  \centering
  \includegraphics[width=\textwidth]{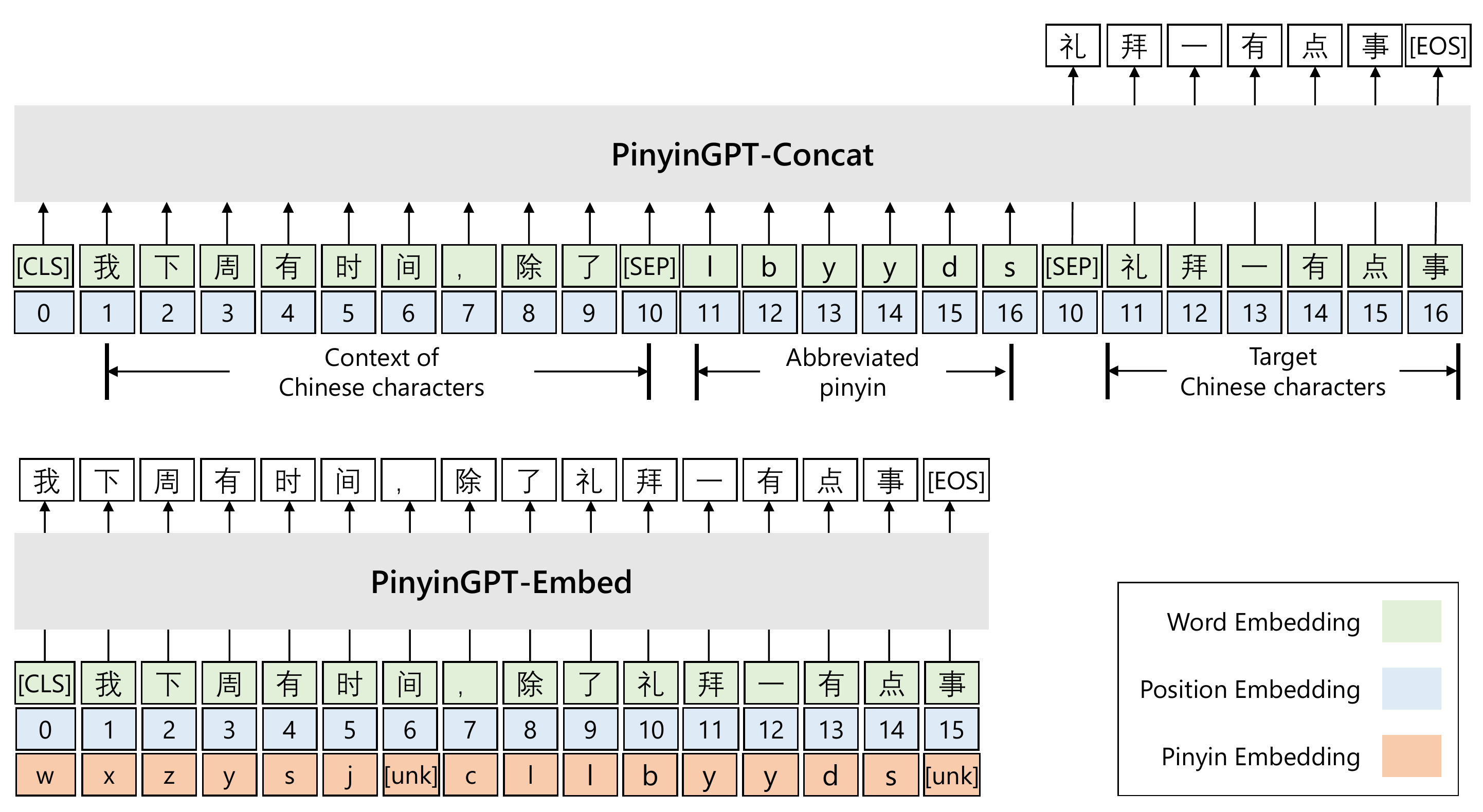}
  \caption{An illustration of the training process of Pinyin-Concat (top) and Pinyin-Embed (bottom), respectively. The example is same as the instance of s2 from Table \ref{tab:task-definition-example}. }
  \label{fig:method-pinyingpt}
\end{figure*}

\subsection{Incorporating Pinyin Context}
\label{sec:pinyin-context}
We explore two simple ways to incorporate pinyin information and build two models correspondingly.
The first model uses pinyin information horizontally by concatenating pinyin input to the context of characters.
The second model incorporates pinyin information vertically by adding a pinyin embedding layer at the bottom of GPT. 

\paragraph{PinyinGPT-Concat}
\label{sec:concatenation}
In this model, we append a pinyin sequence to the context of Chinese characters.
In the inference stage, the input has the form of $\mathbf{x} = [w_1,\dots,w_n,\texttt{[SEP]},p_{n+1},\dots,p_{n+k},\allowbreak\texttt{[SEP]} ]$, 
where \texttt{[SEP]} is a special token to separate text and pinyin.
The model largely follows the architecture of the standard GPT.
Since there is one-one relationship between pinyin tokens and generated Chinese characters (i.e., the pronunciation of $w_{n+j}$ is $p_{n+j}$), we adjust the absolute positions of the characters to be generated. We assign the position of $p_{n+j}$ to $w_{n+j}$, expecting the model to learn the alignments between pinyin and target characters.\footnote{On abbreviated pinyin, this strategy could bring 0.3 points in terms of P@5.}
We further expand the vocabulary of the word embedding layer by adding pinyin tokens. 

In the training stage, given an training instance of $[w_1,\dots,w_n,\texttt{[SEP]},p_{n+1},\dots,p_{n+k},\allowbreak\texttt{[SEP]} ,w_{n+1},\dots,w_{n+k}]$, the model is trained to minimize the following loss function, where $\mathbf{w}_{<n+j}$ stands for the characters before $w_{n+j}$ and $\mathbf{p} = [p_{n+1},\dots,p_{n+k}]$.
\begin{equation}
\mathcal{L_\text{concat}} = - \sum_{j=1}^k \text{log} \ p(w_{n+j}|\mathbf{w}_{<n+j}, \mathbf{p})
\end{equation}

%


\paragraph{PinyinGPT-Embed}
\label{sec:composition}
The original GPT model includes a word embedding layer and a position embedding layer.
In this model, we add a pinyin embedding layer. 
The basic idea is to provide the model with the pinyin of the character to be generated next.
Specifically, the embedding of each character is the sum of the  token embedding of the current character, the position embedding of the current character and the pinyin embedding of the next character. 
When a word~(e.g., numbers, punctuations and symbols) has no corresponding pinyin, we use a special token \texttt{[unk]} to represent it instead.
The training process is similar with the standard GPT, as shown in Figure \ref{fig:method-pinyingpt}.
The loss function is given as follows.
\begin{equation}
\mathcal{L_\text{embed}} = - \sum_{j=1}^{n+k} \text{log} \ p(w_j|\mathbf{w}_{<j}, \mathbf{p}_{<j+1})
\end{equation}
In the inference stage, we transform the input sequence to the same format.




\subsection{Pinyin-Constrained Training}
\label{sec:training}
We describe training details in this subsection.
In standard GPT, the loss function is computed over the whole vocabulary.
However, this is suboptimal for pinyin input method because the major challenge in the inference stage is how to select the best one from characters pronounced with the same pinyin (as described in the end of section \ref{sec:gpt}).
This leads to inconsistency between training and inference stages.
Therefore, in the training stage, the probability of a character is calculated over characters pronounced with the same pinyin, which is formulated as follows.

\begin{equation}
    p(w_i) = \frac{\exp{(g(w_i))}}{\sum_{w_j\in \mathcal{V}_{p_i}}\exp{(g(w_j))}},
\end{equation}
where  $\mathcal{V}_{p_i}$ is the set of Chinese characters whose pinyin is $p_i$ and $g$ is the logit before the softmax layer.

\section{Experiment}

In this section, we show the results 
on  pinyin input method over the two settings (i.e., perfect pinyin and abbreviated pinyin).

\subsection{Settings}
We describe the two datasets used in the following experiments and the evaluation metric.

\paragraph{PD Dataset} PD dataset~\cite{yang-etal-2012-towards} is a commonly used benchmark dataset for the evaluation of pinyin input method \cite{jia-zhao-2014-joint,zhang-etal-arxiv-tracing,huang-etal-2018-moon,zhang-etal-2019-open}. 
The texts in PD are extracted from the People’s Daily\footnote{\url{http://www.people.com.cn/}} from 1992 to 1998.
It contains 5.04 million segments of consecutive Chinese characters (or Maximum Input Unit in some literature) for training and 2,000 segments for testing.
For each test case, the input pinyin are all perfect pinyin and the context is null.

\paragraph{WD Dataset} Since the PD data includes out-of-date news from 20 years ago and does not support us to study the scenario where the context includes real words, we construct a new dataset called WD.
We use the WuDaoCorpora \cite{YUAN202165} that contains 3TB Chinese corpus collected from 822 million Web pages.
Currently, 200GB of the corpus has been made publicly available~\footnote{\url{https://resource.wudaoai.cn/home}}.
We randomly select 15 domains from WuDaoCorpora.
For each domain, we first use an off-the-shelf Chinese segmentation toolkit \cite{texsmart2020} to segment the documents into sentences.
Then we automatically obtain the perfect pinyin and abbreviated pinyin of characters with pinyin converting tools.
For each sentence, we randomly choose a context with a range from 0-3, 4-9 and 10+ words.
Consecutively, we choose the target to be 1-3, 4-9 or 10+ words, respectively.
It's further required that the target should be continuous characters that each has its own pinyin.
We call each context-target length tuple like (4-9, 10+) as an evaluation configuration.
For each configuration, we sample 2,000 test cases.
In total, there are 9 configurations of 18,000 cases for each domain.  The whole dataset consists of 270,000 examples. 
We investigate extremely long target lengths for the purpose of research that these configurations may not appear in real cases.
All the instances in the WD dataset are only used for evaluation.

\paragraph{Evaluation Metric} 
We use precision at top-$K$ as the evaluation metric, which is widely adopted in the literature
~\cite{jia-zhao-2014-joint,zhang-etal-arxiv-tracing,zhang-etal-2019-open}. 
It measures if the ground truth exists in the top-$K$ generated results.
Some existing works also use keystroke-based metrics~\cite{jia-zhao-2013-kyss,huang-etal-2015-ijcai-input} and human evaluation, which we don't use in this work because the evaluation process is more complex and time-consuming.

\paragraph{Other Settings} 

We train both PinyinGPT models with the training data of GPT (ours).
To preprocess the corpus, we use a public library \textit{pypinyin}\footnote{\url{https://github.com/mozillazg/python-pinyin}} to get the pinyin of Chinese characters.\footnote{
If there are heteronym issues, we further verify them with an online dictionary ZDic (\url{https://www.zdic.net/}).}
We initialize both PinyinGPT models with GPT (ours).
Both models are trained  for 100k steps on 32 GPUs of NVIDIA V100 Tensor Core with a bach size of 25,000.
The learning rate is 5e-5.
We maintain a maximum of 128 tokens for every training example.
We use a probability of 50\% to sample a target sequence with less than 5 words, otherwise we randomly sample a target sequence with 6 to 25 words. 
During inference stage, we use beam search with a beam size of 16 for text generation.

\subsection{Results on Perfect Pinyin}
 We report results on the PD dataset~\cite{yang-etal-2012-towards}.
We use pinyin-constraint training in all configurations and train PinyinGPT models with different pinyin vocabularies for perfect pinyin and abbreviated pinyin, respectively.
We compare with the following baselines. 
\begin{itemize}
    \item Google IME is a commercial Chinese IME which provides a debuggable API.
    \item On-OMWA~\cite{zhang-etal-arxiv-tracing} is an online model for word acquisition which adaptively learns new words for Chinese IME. 
    \item On-P2C~\cite{zhang-etal-2019-open} is a neural pinyin-to-Chinese character conversion model, which is augmented by an online updated vocabulary to support open vocabulary learning.
\end{itemize}

In Table~\ref{tab:perfect}, the first group (top) shows the results of the aforementioned baselines, which are directly extracted from On-P2C~\cite{zhang-etal-2019-open}.
The bottom group shows the performance of GPT~(public) and GPT~(ours) with 
%
frozen parameters.
We can find that GPT~(public) achieves comparative performance with existing systems in terms of P@5 and P@10.
After being trained with a larger corpus, GPT~(ours) surpasses all the baseline models in terms of all metrics.
It is worth noting that existing baselines are supervised models that are fine-tuned on training instances. 
The results demonstrate the effectiveness of GPT models pretrained on vast amount of texts.

\begin{table}[t]\centering
\begin{tabular}{lrrrr}\toprule
Model&P@1 &P@5 &P@10 \\\midrule
Google IME  & 70.90 & 78.30 &82.30 \\
On-OMWA  & 64.40 &72.90 &77.90 \\
On-P2C  & 71.30 &80.50 &81.30\\\midrule
GPT (public)  &  67.35& 79.95& 81.60\\
GPT (ours)  & \textbf{73.15} & \textbf{84.10} & \textbf{85.45}  \\
\bottomrule
\end{tabular}
\caption{Comparison with different methods over PD using perfect pinyin. Each score is averaged over all the domains and context-target length configurations.}
\label{tab:perfect}
\end{table}

\subsection{Results on Abbreviated Pinyin}

\begin{table*}[t]
\centering
\begin{tabular}{lcrrrrrrrrr}\toprule
\multirow{3}{*}{Model} &\multirow{2}{*}{Fix GPT} & &\multicolumn{3}{c}{Perfect Pinyin} & &\multicolumn{3}{c}{Abbreviated Pinyin} \\\cmidrule{4-6}\cmidrule{8-10}
& Parameters& &P@1 &P@5 &P@10 & &P@1 &P@5 &P@10 \\\midrule
GPT (public) & & &76.55 &87.07 &88.58 & &22.22 &29.99 &31.48 \\
GPT (ours) & & &80.22 &90.20 &91.09 & &26.90 &35.56 &37.03 \\\midrule
PinyinGPT-Embed &Y & &72.41 &83.44 &84.78 & &26.95 &35.56 &37.06 \\
PinyinGPT-Embed &N & &69.34 &81.54 &82.99 & &23.73 &31.80 &33.33 \\
PinyinGPT-Concat &Y & &\textbf{80.24} &90.21 &91.10 & &26.91 &35.56 &37.03 \\
PinyinGPT-Concat &N & &78.12 &\textbf{90.38} &\textbf{92.06} & &\textbf{27.75} &\textbf{40.66} &\textbf{44.20} \\
\bottomrule
\end{tabular}
\caption{Overall results on WD dataset
for perfect pinyin and  abbreviated pinyin, respectively.}
\label{tab:wd}
\end{table*}

\begin{table*}[t]
\centering
\small
\begin{tabular}{llrrrrrrrrrrrrr}\toprule
 &\multirow{3}{*}{Model}& &\multicolumn{3}{c}{1-3} & &\multicolumn{3}{c}{4-9} & &\multicolumn{3}{c}{10+} \\\cmidrule{4-6}\cmidrule{8-10}\cmidrule{12-14}
& & &P@1 &P@5 &P@10 & &P@1 &P@5 &P@10 & &P@1 &P@5 &P@10 \\\midrule
\multirow{2}{*}{0-3} &GPT~(ours) & &30.11 &42.27 &45.25 & &13.33 &18.24 &18.99 & &4.16 &5.86 &6.00 \\
&PinyinGPT-Concat & &\textbf{31.72} &\textbf{48.09} &\textbf{53.94} &\textbf{} &\textbf{15.21} &\textbf{24.39} &\textbf{26.94} &\textbf{} &\textbf{5.58} &\textbf{9.22} &\textbf{10.09} \\\midrule
\multirow{2}{*}{4-9} &GPT~(ours) & &49.83 &65.03 &67.96 & &25.53 &34.48 &35.89 & &9.38 &12.70 &13.03 \\
&PinyinGPT-Concat & &\textbf{50.78} &\textbf{70.11} &\textbf{75.58} &\textbf{} &\textbf{26.44} &\textbf{41.51} &\textbf{45.52} &\textbf{} &\textbf{10.20} &\textbf{17.02} &\textbf{18.80} \\\midrule
\multirow{2}{*}{10+} &GPT~(ours) & &59.39 &75.00 &77.60 & &\textbf{35.42} &46.32 &47.94 & &\textbf{14.96} &20.11 &20.63 \\
&PinyinGPT-Concat & &\textbf{59.89} &\textbf{78.81} &\textbf{83.33} &\textbf{} &34.99 &\textbf{51.99} &\textbf{56.62} &\textbf{} &14.93 &\textbf{24.78} &\textbf{27.03} \\
\bottomrule
\end{tabular}
\caption{Results of different context-target configurations over WD for abbreviated pinyin. The first column and top row stand for context length range and  target length range, respectively. }
\label{tab:length}
\end{table*}

In this section, we report results for both perfect pinyin and abbreviated pinyin on WD. 

\begin{figure*}[t]
  \centering
  \includegraphics[width=\textwidth]{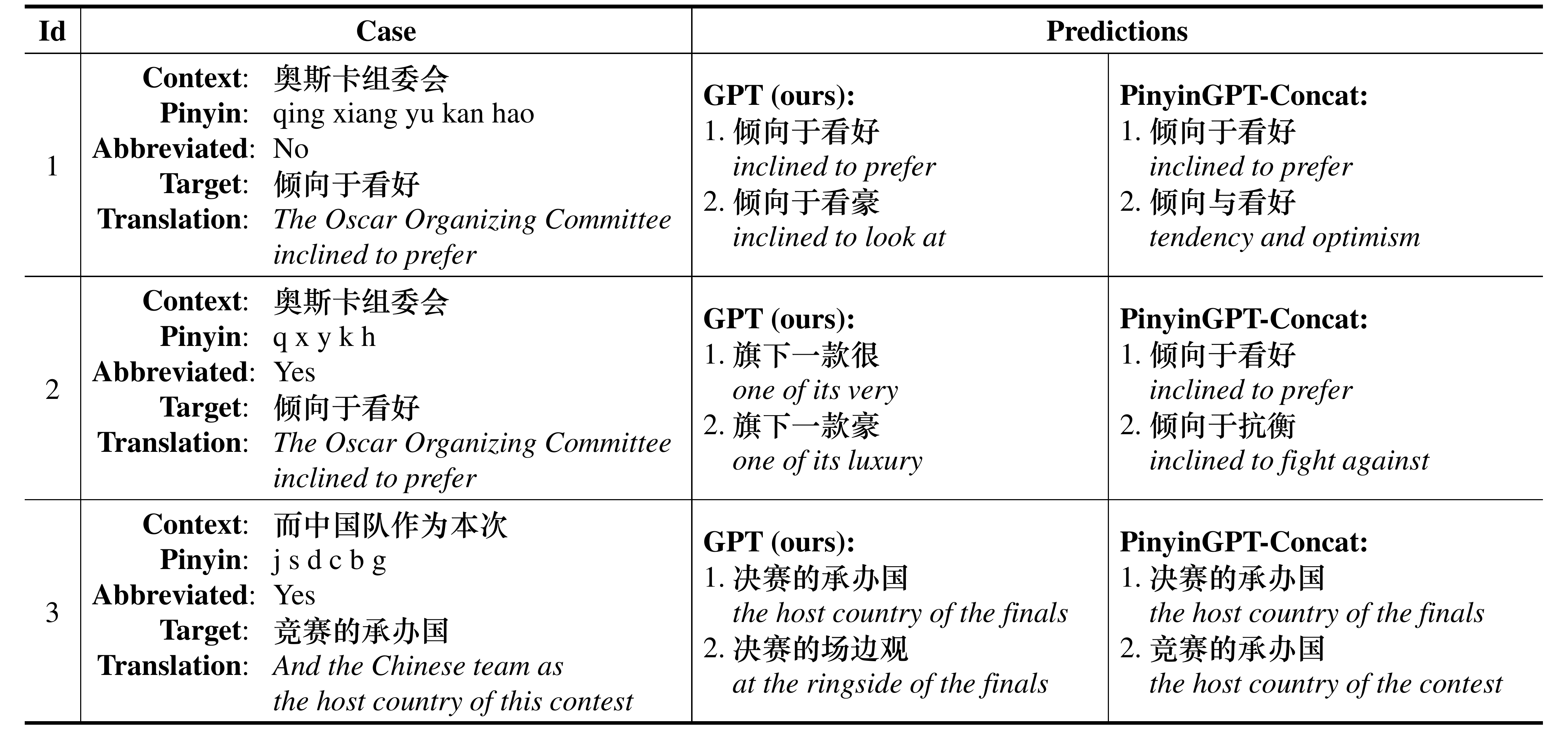}
  \caption{Case study for GPT~(ours) and PinyinGPT-Concat in both perfect pinyin and abbreviated pinyin.}
  \label{fig:case-pinyingpt}
\end{figure*}

In Table~\ref{tab:wd}, we list the overall experiment results of two GPT baselines as well as our PinyinGPT models.
We have several findings based on the results.
First, from each row, we can see that there is a drastic performance drop for all models.
The reason is that each abbreviated pinyin can be mapped to a large amount of candidate characters, so that the problem is more challenging compared to perfect pinyin.
We also believe that the evaluation metric of P@1 might be too strict for abbreviated pinyin because sometimes the top predictions might be correct (as reflected in Figure \ref{fig:case-pinyingpt}) even though they may be different from the ground truth.
Second, adding pinyin information to GPT obtains limited improvement on perfect pinyin, but boosts the abbreviated setting by 5 points on P@5 and 7 points on P@10, respectively.
Third, concatenating pinyin context horizontally is better than adding pinyin embedding vertically.
Last, fine-tuning all the parameters performs better than keeping the parameters of GPT fixed.

\subsection{Model Analysis: Ablation Study}
\begin{figure}[!ht]
\centering
\includegraphics[width=\linewidth]{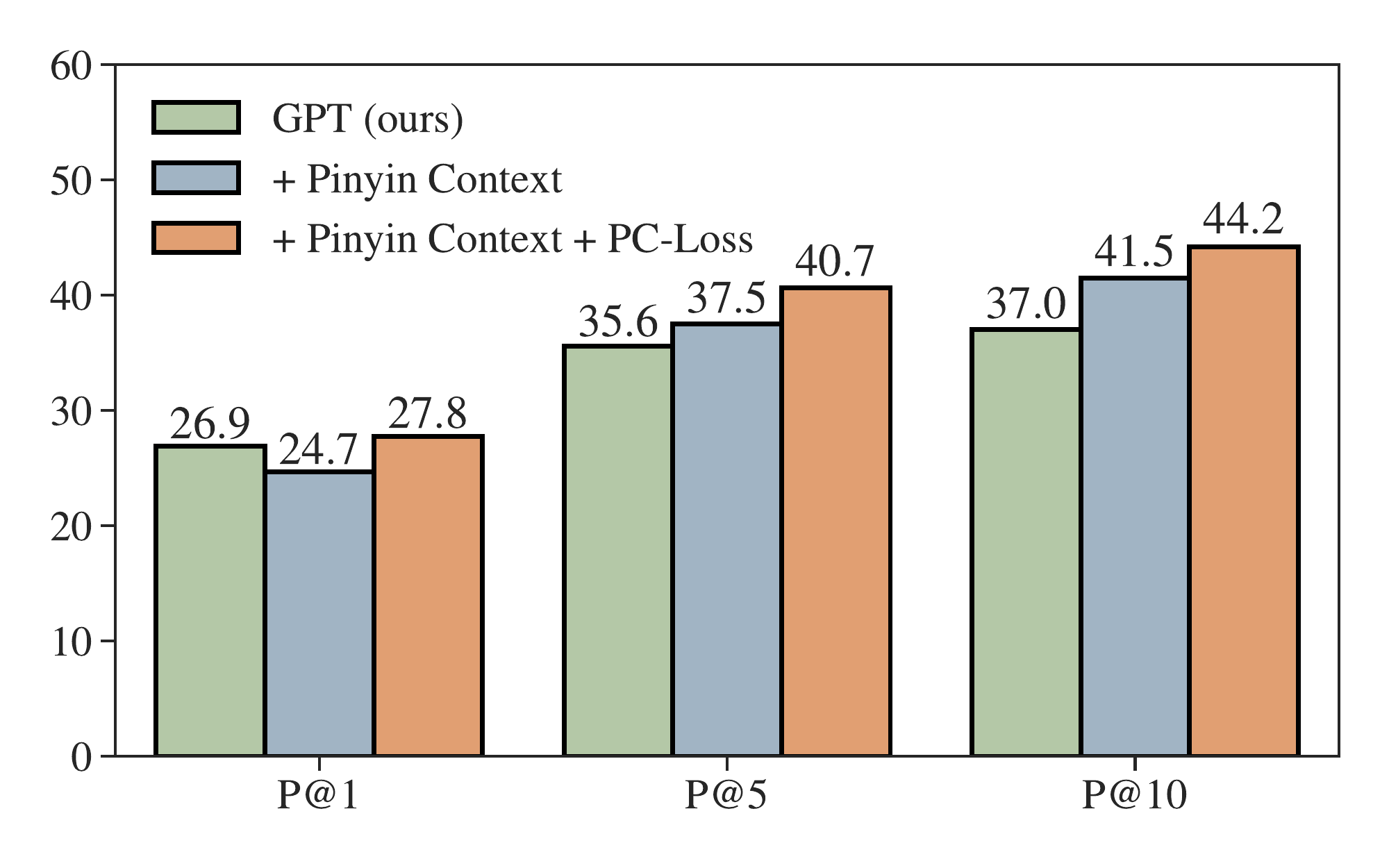}
\caption{Ablation study for concatenating pinyin context and pinyin-constrained training.}
\label{fig:ablation}
\end{figure}

In this section, we conduct experiments to understand the importance of pinyin context and pinyin-constrained training.
Results are given in Figure~\ref{fig:ablation}.
The baseline model is GPT~(ours).
The model \emph{+ Pinyin Context} means that we concatenate pinyin context (i.e., PinyinGPT-Concat) and learn over the whole vocabulary.
The model \emph{+ Pinyin Context + PC-LOSS} means that we use both pinyin context and  pinyin-constrained training.
The figure shows that taking pinyin as extra context works well to improve results in terms of P@5 and P@10.
When the two components are adopted, the performance is further improved.

\subsection{Model Analysis: Context-Target Length}
To analyze how context length and target length affect performance, we aggregate experiment results to form a matrix of accuracy for each configuration in Table~\ref{tab:length}.
Each score is averaged over all the domains. 
From each column, we can see that longer context benefits both GPT and our model in pinyin input method, which verifies the power of context understanding ability of GPT models.
An interesting finding is that, when the context is long enough (e.g., 10+), adding pinyin does not help improve the P@1.

\subsection{Model Analysis: Case Study}

We list three cases in Figure~\ref{fig:case-pinyingpt} to compare model outputs produced by GPT~(ours) and PinyinGPT-Concat.
The first case shows that, given perfect pinyin as the input, both GPT~(ours) and PinyinGPT-Concat make the correct predictions.
In the second case, abbreviated pinyin is given as the input. 
PinyinGPT-Concat makes the correct prediction while the prediction of GPT~(ours) does not fit to the context well.
In Case 3, even if PinyinGPT-Concat ranks the ground truth as the second best, the top 1 prediction still makes much sense and fit well with the context.
In all cases, GPT~(ours) usually generate predictions which are grammatically sound but semantically inappropriate.

\subsection{Model Analysis:  Domains}
In this subsection, we analyze how performance differs with respect to domains.
We put the full table over all domains in the Appendix and sample six domains for illustration in Table~\ref{tab:domain-sample}. 
The table shows that PinyinGPT-Concat achieves consistent improvement over GPT on all domains.
We also find that the absolute scores vary a lot across domains.
This reflects different predictability for texts on different domains. 
For example, the P@10 score of the  \emph{Culture}  domain is 16 points lower than the \emph{Medical} domain.
In the Medical domain, the texts contain plenty of descriptions of symptoms and instructions of medicines, which are somehow canonically used.
While in the Culture domain, the texts are less constrained and have more variations.

\begin{table*}[!htp]
\centering
\small
\begin{tabular}{lrrrrrrrrrrrrr}\toprule
\multirow{3}{*}{Model} & &\multicolumn{3}{c}{Games} & &\multicolumn{3}{c}{Culture} & &\multicolumn{3}{c}{Sports} \\\cmidrule{3-5}\cmidrule{7-9}\cmidrule{11-13}
& &P@1 &P@5 &P@10 & &P@1 &P@5 &P@10 & &P@1 &P@5 &P@10 \\\midrule
GPT (ours) & &24.04 &32.78 &34.23 & &21.86 &29.33 &30.94 & &28.54 &37.13 &38.69 \\
PinyinGPT-Concat & &\textbf{25.78} &\textbf{38.26} &\textbf{41.89} &\textbf{} &\textbf{22.10} &\textbf{33.33} &\textbf{36.72} & &\textbf{29.81} &\textbf{43.56} &\textbf{46.95} \\
& & & & & & & & & & & & \\
& &\multicolumn{3}{c}{Real Estate} & &\multicolumn{3}{c}{Medical} & &\multicolumn{3}{c}{Finance} \\\cmidrule{3-5}\cmidrule{7-9}\cmidrule{11-13}
& &P@1 &P@5 &P@10 & &P@1 &P@5 &P@10 & &P@1 &P@5 &P@10 \\\midrule
GPT (ours) & &26.53 &35.27 &36.74 & &33.59 &43.54 &44.93 & &29.00 &37.24 &38.47 \\
PinyinGPT-Concat & &\textbf{27.28} &\textbf{40.16} &\textbf{43.86} &\textbf{} &\textbf{34.76} &\textbf{49.28} &\textbf{52.56} &\textbf{} &\textbf{29.17} &\textbf{42.17} &\textbf{45.52} \\
\bottomrule
\end{tabular}
\caption{Results of 6 sample domains over WD using abbreviated pinyin. Each score is averaged over all the context-target length configurations. The table of all 15 domains is attached in the Appendix.}
\label{tab:domain-sample}
\end{table*} 

\subsection{Model Analysis: Accuracy versus Latency}
Considering pinyin input method requires both accuracy and efficiency, we further train a 6-layer GPT to investigate the trade-off.
Our 6-layer GPT is directly truncated and initialized from the 12-layer GPT and is continually trained for 50k steps with the same configuration of 12-layer GPT.

The evaluation is conducted over the 9 configurations of context-target length and averaged across all domains.
Specifically, each configuration is inferred using a data center GPU NVIDIA V100 Tensor Core, and the GPU is fully occupied by one model.
The beam size is set to be 16.
We report the average inference time in millisecond as well as accuracy in terms of P@$K$ of PinyinGPT-Concat. 
Table~\ref{tab:trade-off-4-9} is the result for the configuration (4-9, 4-9).
The table shows that the inference time of the model with 6-layer transformer is almost 30\% faster than that with 12-layer.
However, the performance of the 6-layer model drops consistently in the abbreviated setting.
We also list the experiment results for all configurations in the Appendix.
We recommend readers to select models in a more cost-effective way based on their requirements.

\begin{table}[!htp]
\centering
\begin{tabular}{lcc}\toprule
Model &Time (ms) &P@5 \\\midrule
GPT~(ours, 6L) & 94 &27.45 \\
GPT~(ours, 12L) & 142 &34.48 \\
PinyinGPT-Concat~(6L) &94 &32.70 \\
PinyinGPT-Concat~(12L) &145 &41.51 \\
\bottomrule
\end{tabular}
\caption{Average inference time for one instance and the overall P@5 for the configuration of (4-9, 4-9). 
}
\label{tab:trade-off-4-9}
\end{table}

\section{Related work}

We describe related works on pinyin input method and pinyin-enhanced pretrained  models here.

\paragraph{Pinyin Input Method} We describe existing works based on whether the input pinyin is perfect or abbreviated.
A majority of existing works focus on {perfect} pinyin.
Traditional models are typically based on statistical language models~\cite{chen-lee-2000-new} and statistical machine translation~\cite{yang-etal-2012-towards}.
Recent works are usually built with neural network. For example, Moon IME~\cite{huang-etal-2018-moon} integrates  attention-based neural network and an information retrieval module. 
\citet{zhang-etal-2019-open} improves an LSTM-based encoder-decoder model with online vocabulary adaptation.
For abbreviated pinyin, 
CoCAT~\cite{huang-etal-2015-ijcai-input} uses machine translation technology to reduce the number of the typing letters. 
\citet{huang-zhao-2018-chinese} propose an LSTM-based encoder-decoder approach with the concatenation of context words and abbreviated pinyin as input. Our work differs from existing works in that we are the first one to exploit GPT and verify the pros and cons of GPT in different situations.
In addition, there are some works handling pinyin with typing errors.
\citet{chen-lee-2000-new} investigate a typing model which handles spelling correction in sentence-based pinyin input method.
CHIME~\cite{zheng-etal-ijcai-2011-chime} is a error-tolerant Chinese pinyin input method. It finds similar pinyin which will be further ranked with Chinese specific features.
\citet{jia-zhao-2014-joint} propose a joint graph model to globally optimize the tasks of pinyin input method and typo correction.
We leave error-tolerant pinyin input method as a future work.



\paragraph{Pinyin-enhanced Pretrained Models}
Our methodology also relates to pretrained models that use pinyin information.
\citet{sun-etal-2021-chinesebert} propose a general-purpose Chinese BERT with new embedding layers to inject pinyin and glyph information of characters.
There are also task-specific BERT models, especially for the task of grammatical error correction since an important type of error is caused by characters pronounced with the same pinyin.
\citet{zhang-etal-2021-correcting} add a pinyin embedding layer and learns to predict characters from similarly pronounced candidates.
PLOME~\cite{liu-etal-2021-plome} add two embedding layers implemented with two GRU networks to inject both pinyin and shape of characters, respectively.
\citet{xu-etal-2021-read} add a hierarchical encoder to inject the pinyin letters at character and sentence levels, and add a ResNet encoder to use graphic features of character image.  

\section{Conclusion}

In this paper, we explore how to adapt pretrained Chinese GPT to pinyin input method.
To begin with, we find that a frozen GPT with decoding conditioned on pinyin can reach state-of-the-art performance on perfect pinyin. 
However, in abbreviated setting, the performance drops by a large gap.
Through our experiments, we find that both context enrichment with pinyin and pinyin-constrained training improve the performance.
In the future, we would like to investigate more challenging settings including error-tolerant pinyin input method and mixtures of perfect pinyin and abbreviated pinyin.

\bibliography{anthology,custom}
\bibliographystyle{acl_natbib}

\appendix

\begin{table*}[!htp]\centering
\small
\begin{tabular}{lrrrrrrrrrrrrr}\toprule
Model & &Top1 &Top5 &Top10 & &Top1 &Top5 &Top10 & &Top1 &Top5 &Top10 \\\midrule
& &\multicolumn{3}{c}{Entertainment} & &\multicolumn{3}{c}{Automobile} & &\multicolumn{3}{c}{Technology} \\\cmidrule{3-5}\cmidrule{7-9}\cmidrule{11-13}
GPT~(ours) & &26.84 &35.97 &37.73 & &27.84 &36.56 &38.03 & &26.01 &34.48 &35.86 \\
PinyinGPT-Concat & &\textbf{28.74} &\textbf{41.68} &\textbf{45.48} &\textbf{} &\textbf{28.74} &\textbf{41.55} &\textbf{45.28} &\textbf{} &\textbf{26.82} &\textbf{40.17} &\textbf{43.65} \\
\midrule
& &\multicolumn{3}{c}{Education} & &\multicolumn{3}{c}{Agriculture} & &\multicolumn{3}{c}{Economy} \\\cmidrule{3-5}\cmidrule{7-9}\cmidrule{11-13}
GPT~(ours) & &27.31 &36.71 &38.28 & &26.57 &35.08 &36.59 & &27.93 &36.04 &37.20 \\
PinyinGPT-Concat & &\textbf{27.65} &\textbf{41.17} &\textbf{44.87} & &\textbf{27.27} &\textbf{39.73} &\textbf{43.17} &\textbf{} &\textbf{28.47} &\textbf{40.99} &\textbf{44.53} \\\midrule
& &\multicolumn{3}{c}{Games} & &\multicolumn{3}{c}{Culture} & &\multicolumn{3}{c}{Sports} \\\cmidrule{3-5}\cmidrule{7-9}\cmidrule{11-13}
GPT~(ours) & &24.04 &32.78 &34.23 & &21.86 &29.33 &30.94 & &28.54 &37.13 &38.69 \\
PinyinGPT-Concat & &\textbf{25.78} &\textbf{38.26} &\textbf{41.89} &\textbf{} &\textbf{22.10} &\textbf{33.33} &\textbf{36.72} & &\textbf{29.81} &\textbf{43.56} &\textbf{46.95} \\\midrule
& &\multicolumn{3}{c}{International} & &\multicolumn{3}{c}{Society} & &\multicolumn{3}{c}{Military} \\\cmidrule{3-5}\cmidrule{7-9}\cmidrule{11-13}
GPT~(ours) & &26.42 &34.82 &36.24 & &26.57 &36.15 &37.78 & &24.46 &32.26 &33.75 \\
PinyinGPT-Concat & &\textbf{27.49} &\textbf{40.16} &\textbf{43.66} &\textbf{} &\textbf{27.34} &\textbf{40.94} &\textbf{44.89} &\textbf{} &\textbf{24.82} &\textbf{36.73} &\textbf{40.03} \\\midrule
& &\multicolumn{3}{c}{Real Estate} & &\multicolumn{3}{c}{Medical} & &\multicolumn{3}{c}{Finance} \\\cmidrule{3-5}\cmidrule{7-9}\cmidrule{11-13}
GPT~(ours) & &26.53 &35.27 &36.74 & &33.59 &43.54 &44.93 & &29.00 &37.24 &38.47 \\
PinyinGPT-Concat & &\textbf{27.28} &\textbf{40.16} &\textbf{43.86} &\textbf{} &\textbf{34.76} &\textbf{49.28} &\textbf{52.56} &\textbf{} &\textbf{29.17} &\textbf{42.17} &\textbf{45.52} \\
\bottomrule
\end{tabular}
\caption{Results of different domains over WD using abbreviated pinyin. Each score is averaged over all the context-target length configurations.}
\label{tab:domain}
\end{table*}

\begin{table*}[!htp]\centering
\scriptsize
\begin{tabular}{llrrrrrrrrrrrrrrrr}\toprule
&\multirow{3}{*}{Models} & &\multicolumn{4}{c}{1-3} & &\multicolumn{4}{c}{4-9} & &\multicolumn{4}{c}{10+} \\\cmidrule{4-7}\cmidrule{9-12}\cmidrule{14-17}
& & &T &P@1 &P@5 &P@10 & &T &P@1 &P@5 &P@10 & &T &P@1 &P@5 &P@10 \\\midrule
\multirow{4}{*}{0-3} &GPT~(ours, 6L) & &38 &26.74 &38.45 &41.50 & &98 &10.46 &14.41 &15.19 & &201 &2.72 &3.70 &3.85 \\
&GPT~(ours, 12L) & &58 &30.11 &42.27 &45.25 & &148 &13.33 &18.24 &18.99 & &303 &4.16 &5.86 &6.00 \\
&PinyinGPT-Concat~(6L) & &40 &29.17 &45.17 &50.73 & &98 &11.92 &19.55 &21.84 & &197 &3.20 &5.67 &6.22 \\
&PinyinGPT-Concat~(12L) & &61 &31.72 &48.09 &53.94 & &148 &15.21 &24.39 &26.94 & &305 &5.58 &9.22 &10.09 \\\midrule
\multirow{4}{*}{4-9} &GPT~(ours, 6L) & &38 &44.02 &59.02 &62.32 & &94 &20.02 &27.45 &28.76 & &198 &5.72 &8.05 &8.31 \\
&GPT~(ours, 12L) & &57 &49.83 &65.03 &67.96 & &142 &25.53 &34.48 &35.89 & &301 &9.38 &12.70 &13.03 \\
&PinyinGPT-Concat~(6L) & &38 &45.66 &65.08 &70.56 & &94 &20.25 &32.70 &36.14 & &192 &5.98 &10.23 &11.29 \\
&PinyinGPT-Concat~(12L) & &58 &50.78 &70.11 &75.58 & &145 &26.44 &41.51 &45.52 & &298 &10.20 &17.02 &18.80 \\\midrule
\multirow{4}{*}{10+} &GPT~(ours, 6L) & &42 &54.38 &69.94 &72.92 & &99 &28.81 &38.98 &40.41 & &198 &10.32 &14.18 &14.64 \\
&GPT~(ours, 12L) & &64 &59.39 &75.00 &77.60 & &149 &35.42 &46.32 &47.94 & &301 &14.96 &20.11 &20.63 \\
&PinyinGPT-Concat~(6L) & &43 &53.91 &73.21 &78.14 & &98 &27.21 &42.36 &46.45 & &198 &9.15 &15.49 &17.05 \\
&PinyinGPT-Concat~(12L) & &66 &59.89 &78.81 &83.33 & &154 &34.99 &51.99 &56.62 & &306 &14.93 &24.78 &27.03 \\
\bottomrule
\end{tabular}
\caption{Experiment results for different configurations over WD using abbreviated pinyin, each score is averaged over all the domains. The first column is the context length while the first row is the target length. The field \emph{T} is the average inference time in millisecond.}
\label{tab:trade-off}
\end{table*}

\end{CJK*}

\end{document}